\pdfoutput=1

\documentclass[11pt]{article}

\usepackage[preprint]{acl}

\usepackage{times}
\usepackage{latexsym}
\usepackage{url}
\usepackage{svg}
\usepackage{inconsolata}
\usepackage{times}
\usepackage{colortbl}
\usepackage{latexsym}
\usepackage{amsmath}
\usepackage{amssymb}
\usepackage{array}
\usepackage{pifont}
\usepackage{microtype}
\usepackage{tabularx}
\usepackage{adjustbox}
\usepackage{enumitem}
\usepackage{multirow}
\usepackage{dsfont}
\usepackage{booktabs}
\usepackage{algorithm2e}
\usepackage{MnSymbol}
\usepackage{scalerel}
\usepackage{mathrsfs}
\usepackage{pifont}
\usepackage{multirow}
\usepackage{graphicx}
\usepackage{changes}
\usepackage{xcolor,soul}
\usepackage{makecell}
\usepackage{ulem}
\usepackage{graphicx,calc}
\usepackage{bbm}
\usepackage{lipsum}
\usepackage{bbding}
\usepackage[caption=false]{subfig}
\usepackage{amsmath}

\usepackage{listings}

\usepackage{relsize}
\definecolor{lblue}{HTML}{A6CEE3}
\definecolor{lgreen}{HTML}{B2DF8A}
\definecolor{lred}{HTML}{FB9A99}
\definecolor{lorange}{HTML}{FDBF6F}
\definecolor{mblue}{HTML}{80B1D3}
\definecolor{mgreen}{HTML}{B3DE69}
\definecolor{mred}{HTML}{FB8072}
\definecolor{morange}{HTML}{FDB462}
\definecolor{blue}{HTML}{1F78B4}
\definecolor{green}{HTML}{33A02C}
\definecolor{red}{HTML}{E31A1C}
\definecolor{orange}{HTML}{FF7F00}
\definecolor{dblue}{HTML}{08519C}
\definecolor{dgreen}{HTML}{006D2C}
\definecolor{dorange}{HTML}{EC7014}

\usepackage[T1]{fontenc}

\usepackage[utf8]{inputenc}

\usepackage{microtype}

\usepackage{inconsolata}

\newcommand{\LibraryName}[0]{{\textbf{\texttt{pyvene}}}}
\newcommand{\LibraryUrl}[0]{{\url{https://github.com/stanfordnlp/pyvene}}}

\newcommand{\authcomma}{\textmd{,}\hskip0.5em}
\newcommand{\Secref}[1]{Section~\ref{#1}}

\newcommand{\Figref}[1]{Figure~\ref{#1}}

\title{%
\LibraryName{}: A Library for Understanding and Improving PyTorch Models\\ via Interventions}

\author{Zhengxuan Wu$^{\dag}$\authcomma Atticus Geiger$^{\ddag}$\authcomma Aryaman Arora$^{\dag}$\authcomma Jing Huang$^{\dag}$\authcomma Zheng Wang$^{\dag}$\authcomma\\
\textbf{Noah D. Goodman}$^{\dag}$\authcomma\textbf{Christopher D.~Manning}$^{\dag}$\authcomma\textbf{Christopher Potts}$^{\dag}$\\
$^\dag$Stanford University \quad$^\ddag$Pr(Ai)$^2$R Group\\
\texttt{\{wuzhengx,atticusg,aryamana,hij,peterwz,ngd,manning,cgpotts\}@stanford.edu}\\
}

\begin{document}
\maketitle
\begin{abstract}
Interventions on model-internal states are fundamental operations in many areas of AI, including model editing, steering, robustness, and interpretability. To facilitate such research, we introduce \LibraryName{}, an open-source Python library that supports customizable interventions on a range of different PyTorch modules. \LibraryName{} supports complex intervention schemes with an intuitive configuration format, and its interventions can be static or include trainable parameters. We show how \LibraryName{} provides a unified and extensible framework for performing interventions on neural models and sharing the intervened upon models with others. We illustrate the power of the library via interpretability analyses using causal abstraction and knowledge localization. We publish our library through Python Package Index (PyPI) and provide code, documentation, and tutorials at \LibraryUrl{}.
\end{abstract}

\begin{figure}[t!]
  \centering
  \includegraphics[width=1.0\linewidth]{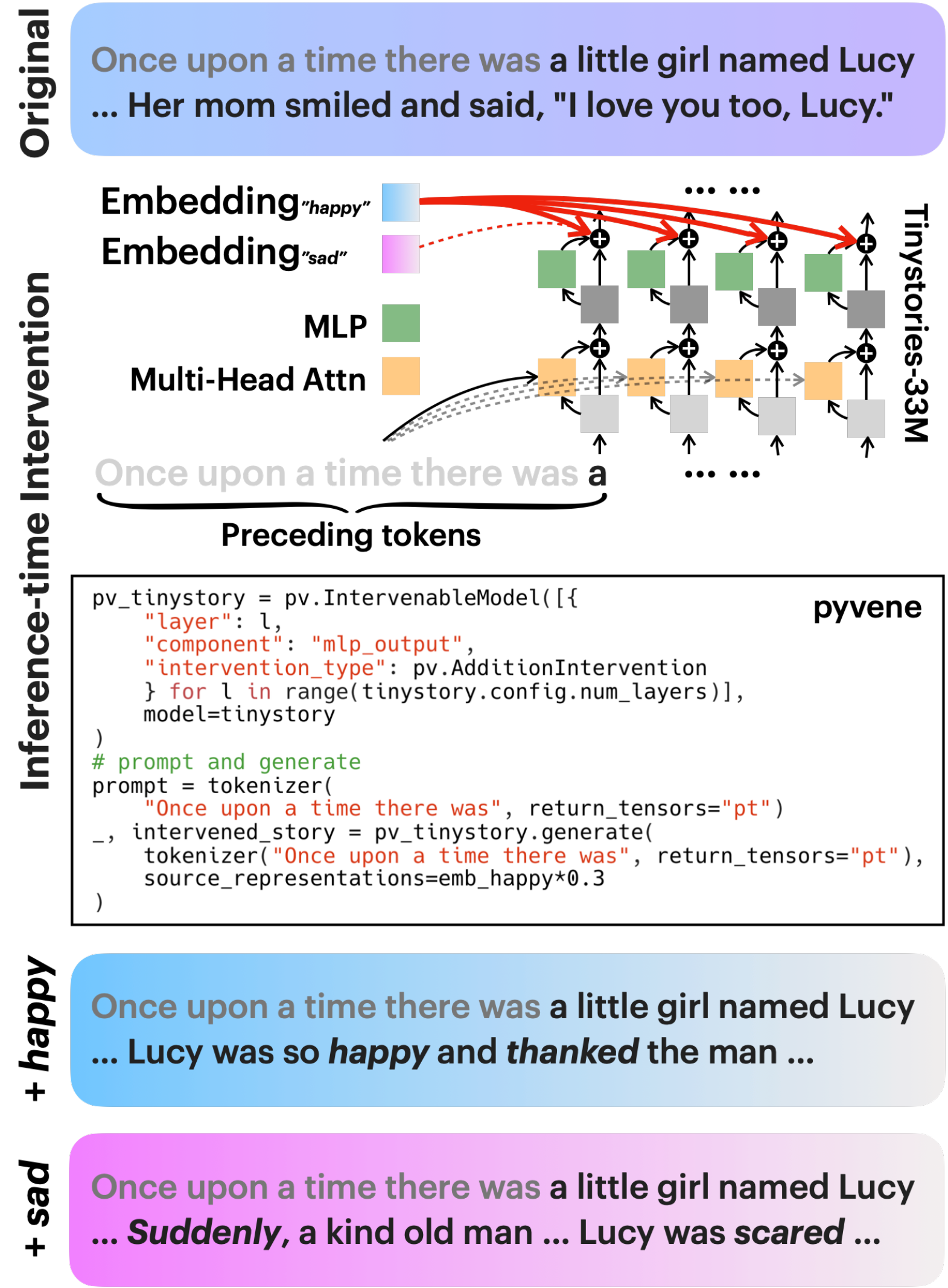}
  \caption{An inference-time intervention~\citep{li2023inferencetime} on \texttt{TinyStories-33M}. The model is prompted with ``Once upon a time there was a'', and is asked to complete the story. We add a static word embedding (for ``happy'' or ``sad'') into the MLP output at each decoding step for all layers with a coefficient of 0.3. \LibraryName{}'s complete implementation is provided. The original and intervened generations use greedy decoding.}
  \label{fig:main}
\end{figure}

\section{Introduction}

When we \emph{intervene} on a neural network, we make an in-place change to its activations, putting the model in a counterfactual state. This fundamental operation has emerged as a powerful tool for both understanding and improving models; interventions of various kinds are key to recent efforts in model robustness~\cite{He2019Parametric}, model editing~\cite{meng2022locating} and steering~\cite{li2023inferencetime}, causal abstraction~\cite{Geiger-etal-2020,Geiger:2021,geiger2023finding,wu-etal-2023-Boundless-DAS} or activation patching~\cite{chan2022causal,wang2022interpretability}, circuit finding~\cite{conmy2023towards,goldowsky2023localizing}, and knowledge tracing~\cite{geva-etal-2023-dissecting}. 

As intervention-based techniques have matured, the need has arisen to run ever more complex interventions on ever larger models. Currently, there is no unified and generic intervention-oriented library to support such research. Existing libraries are often project-based (see implementations for \citealt{wang2022interpretability,geiger2023finding} as examples) that lack extensibility and are hard to maintain and share, and current toolkits focus on single or non-nested interventions (e.g., ablation neurons in a single forward pass) and are often limited to interventions on Transformers \citep{Vaswani-etal:2017} without natively supporting other neural architectures. Some of these existing libraries \citep{baukit,graphpatch,nnsight,mossing2024tdb} can support complex interventions such as exchanging activations across multiple forward passes yet they require sophisticated knowledge and heavy implementations.

To address these limitations, we introduce \LibraryName{},
an open-source Python library that supports customizable interventions on different neural architectures implemented in PyTorch. Different from previous libraries~\cite{baukit,transformerlens,graphpatch,nnsight,mossing2024tdb}, \LibraryName{} is intervention-oriented. It supports complex interventions by manipulating or exchanging activations across multiple model forward runs while allowing these interventions to be shared with a serialization configuration file. Specifically, \LibraryName{} has a number of advantages:

\begin{enumerate}[itemsep=0pt, topsep=4pt]
\item \textbf{Intervention as the primitive.} The intervention is the basic primitive of \LibraryName{}. Interventions are specified with a \texttt{dict}-based format, in contrast to previous approaches where interventions are expressed as code and executed during runtime~\cite{baukit,graphpatch,nnsight,mossing2024tdb}. All \LibraryName{} intervention schemes and models are serializable objects that can be shared through a public model hub such as HuggingFace.

\item \textbf{Complex intervention schemes.} \LibraryName{}  supports interventions at multiple locations, involving arbitrary subsets of neurons, and interventions can be performed in parallel or in sequence. For generative use of LMs, \LibraryName{} supports interventions at decoding steps. 
Furthermore, activations can easily be collected for probe training.

\item \textbf{Support for recurrent and non-recurrent models.} Existing libraries offer only limited support for recurrent models.
\LibraryName{} supports simple feed-forward networks, Transformers, and recurrent and convolutional neural models.
 
\end{enumerate}

In this paper, we provide two detailed case studies using \LibraryName{} as well: (1)~we fully reproduce \citet{meng2022locating}'s locating factual associations in \texttt{GPT2-XL} (Figure~1 in the original paper) in about 20 lines of code, and (2)~we show intervention and probe training with \LibraryName{} to localize gender in \texttt{Pythia-6.9B}. \LibraryName{} is published through the Python Package Index (PyPI),\footnote{\texttt{pip install pyvene}} and the project site\footnote{\LibraryUrl{}} hosts more than 20 tutorials that cover interventions at different levels of complexity with various model architectures from simple feed-foward models to multi-modal models.

\section{System Design and Architecture}

Two primary components of \LibraryName{} are the \textbf{intervenable configuration}, which outlines which model components will be intervened upon, and the \textbf{intervenable model}, which decorates the original \texttt{torch} model with hooks that allow activations to be collected and overwritten.\footnote{Code snippets provided in the paper can be run on Google Colab at \url{https://colab.research.google.com/github/stanfordnlp/pyvene/blob/main/pyvene_101.ipynb}.} Here is a setup for performing a zero-out intervention (often called a zero ablation;~\citealt{li2023circuit}) on the 10th, 11th, and 12th dimensions of the MLP output for 3rd token embedding of layer 0 in GPT-2:
\begin{lstlisting}
import torch
import pyvene as pv
# built-in helper to get a HuggingFace model
_, tokenizer, gpt2 = pv.create_gpt2()
# create with dict-based config
pv_config = pv.IntervenableConfig({
  "layer": 0,
  "component": "mlp_output",
  "intervention_type": pv.VanillaIntervention})
# initialize model
pv_gpt2 = pv.IntervenableModel(
  pv_config, model=gpt2)
# run an intervened forward pass
intervened_outputs = pv_gpt2(
  # the base input
  base=tokenizer(
    "The capital of Spain is", 
    return_tensors="pt"), 
  # the location to intervene at (3rd token)
  unit_locations={"base": 3},
  # the individual dimensions targetted
  subspaces=[10,11,12],
  # the intervention values
  source_representations=torch.zeros(
    gpt2.config.n_embd)
)
# sharing
pv_gpt2.save("./tmp/", save_to_hf_hub=True)
\end{lstlisting}
The model takes a tensor input \texttt{base} and runs through the model's computation graph modifying activations in place to be other values \texttt{source}. In this code, we specified \texttt{source} in the forward call. When \texttt{source} is a constant, it can alternatively be specified in the \texttt{IntervenableConfig}. To target complete MLP output representations, one simply leaves out the \texttt{subspaces} argument.
The final line of the code block shows how to serialize and share an intervened model remotely through a model hub such as HuggingFace.

\subsection{Interchange Interventions} 

Interchange interventions~\cite[][also known as activation patching]{Geiger-etal-2020,vig2020causal,wang2022interpretability} fix activations to take on the values they would be if a different input were provided. With minor changes to the forward call, we can perform an interchange intervention on GPT-2:
\begin{lstlisting}
# run an interchange intervention 
intervened_outputs = pv_gpt2(
  # the base input
  base=tokenizer(
    "The capital of Spain is", 
    return_tensors = "pt"), 
  # the source input
  sources=tokenizer(
    "The capital of Italy is", 
    return_tensors = "pt"), 
  # the location to intervene at (3rd token)
  unit_locations={"sources->base": 3},
  # the individual dimensions targeted
  subspaces=[10,11,12]
)
\end{lstlisting}
This forward call produces outputs for \texttt{base} but with the activation values for MLP output dimensions 10--12 of token 3 at layer 0 set to those that obtained when the model processes the source. Such interventions are used in interpretability research to test hypotheses about where and how information is stored in model-internal representations.

\subsection{Addition Interventions} 

In the above examples, we replace values in the base with other values (\texttt{VanillaIntervention}). Another common kind of intervention involves updating the base values in a systematic way:
\begin{lstlisting}
noising_config = pv.IntervenableConfig({
  "layer": 0,
  "component": "block_input",
  "intervention_type": pv.AdditionIntervention})
noising_gpt2 = pv.IntervenableModel(
  config, model=gpt2)
intervened_outputs = noising_gpt2(
  base=tokenizer(
    "The Space Needle is in downtown", 
    return_tensors = "pt"), 
  # target the first four tokens for intervention
  unit_locations={"base": [0, 1, 2, 3]},
  source_representations = torch.rand(
    gpt2.config.n_embd, requires_grad=False))
\end{lstlisting}  
As in this example, we add noise to a representation as a basic robustness check. The code above does this, targetting the first four input token embeddings to a Transformer by using \texttt{AdditionIntervention}. This example serves as the building block of causal tracing experiments as in \citealt{meng2022locating}, where we corrupt embedding inputs by adding noise to trace factual associations. Building on top of this, we reproduce \citeauthor{meng2022locating}'s result in \Secref{sec:gpt2xl}. \LibraryName{} allows \texttt{Autograd} on the static representations, so this code could be the basis for training models to be robust to this noising process.

\subsection{Activation Collection Interventions} 

This is a pass-through intervention to collect activations for operations like supervised probe training. Such interventions can be combined with other interventions as well, to support things like causal structural probes~\cite{hewitt2019structural, elazar2020amnesic, lepori2023uncovering}. In the following example, we perform an interchange intervention at layer~8 and then collect activations at layer~10 for the purposes of fitting a probe:
\begin{lstlisting}
# set up a upstream intervention
probe_config = pv.IntervenableConfig({
  "layer": 8,
  "component": "block_output",
  "intervention_type":  pv.VanillaIntervention})
# add downstream collector
probe_config = probe_config.add_intervention({
  "layer": 10,
  "component": "block_output",
  "intervention_type": pv.CollectIntervention})
probe_gpt2 = pv.IntervenableModel(
  probe_config, model=gpt2)
# return the activations for 3rd token
collected_activations = probe_gpt2(
  base=tokenizer(
    "The capital of Spain is", 
    return_tensors="pt"), 
  unit_locations={"sources->base": 3})
\end{lstlisting} 

\subsection{Custom Interventions} 

\LibraryName{} provides a flexible way of adding new intervention types. The following is a simple illustration in which we multiply the original representation by a constant value:
\begin{lstlisting}
# multiply base with a constant
class MultInt(pv.ConstantSourceIntervention):
  def __init__(self, **kwargs):
    super().__init__()
  def forward(self, base, source=None, 
    subspaces=None):
    return base * 0.3
       
pv.IntervenableModel({
  "intervention_type": MultInt}, 
  model=gpt2)
\end{lstlisting} 
The above intervention becomes useful when studying interpretability-driven models such as the  Backpack LMs of \citet{hewitt-etal-2023-backpack}. The sense vectors acquired during pretraining in Backpack LMs have been shown to have a ``multiplication effect'', and so proportionally decreasing sense vectors could effectively steer the model's generation.

\subsection{Trainable Interventions} 

\LibraryName{} interventions can include trainable parameters. %
\texttt{RotatedSpaceIntervention} implements Distributed Alignment Search (DAS; \citealt{geiger2023finding}), \texttt{LowRankRotatedSpaceIntervention}~is~a more efficient version of that model, and \texttt{BoundlessRotatedSpaceIntervention} implements the Boundless DAS variant of \citet{wu-etal-2023-Boundless-DAS}. With these primitives, one can easily train DAS explainers.

In the example below, we show a single gradient update for a DAS training objective that localizes the capital associated with the country in a one-dimensional linear subspace of activations from the Transformer block output (i.e., main residual stream) at the 8th layer by training our intervention module to match the gold counterfactual behavior:

\begin{lstlisting}
das_config = pv.IntervenableConfig({
  "layer": 8,
  "component": "block_output",
  "low_rank_dimension": 1, 
  "intervention_type": 
    pv.LowRankRotatedSpaceIntervention})

das_gpt2 = pv.IntervenableModel(
    das_config, model=gpt2)

last_hidden_state = das_gpt2(
  base=tokenizer(
    "The capital of Spain is", 
    return_tensors="pt"), 
  sources=tokenizer(
    "The capital of Italy is", 
    return_tensors="pt"), 
  unit_locations={"sources->base": 3}
)[-1].last_hidden_state[:,-1]

# gold counterfacutual label as " Rome"
label = tokenizer.encode(
  " Rome", return_tensors="pt")
logits = torch.matmul(
  last_hidden_state, gpt2.wte.weight.t())

m = torch.nn.CrossEntropyLoss()
loss = m(logits, label.view(-1))
loss.backward()
\end{lstlisting}  

\subsection{Training with Interventions}

Interventions can be co-trained with the intervening model for techniques like interchange intervention training (IIT), which induce specific causal structures in neural networks~\cite{geiger-etal-2021-iit}:
\begin{lstlisting}
pv_gpt2 = pv.IntervenableModel({
  "layer": 8}, 
  model=gpt2)
# enable gradients on the model
pv_gpt2.enable_model_gradients()
# run counterfactual forward as usual
\end{lstlisting} 
In the example above, with the supervision signals from the training dataset, we induce causal structures in the residual stream at 8th layer.

\subsection{Multi-Source Parallel Interventions}
In the parallel mode, interventions are applied to the computation graph of the same \texttt{base} example at the same time. We can perform interchange interventions by taking activations from multiple source examples and swapping them into the \texttt{base}'s computation graph:
\begin{lstlisting}
parallel_config = pv.IntervenableConfig([
  {"layer": 3, "component": "block_output"},
  {"layer": 3, "component": "block_output"}],
  # intervene on base at the same time
  mode="parallel")

parallel_gpt2 = pv.IntervenableModel(
  parallel_config, model=gpt2)

base = tokenizer(
  "The capital of Spain is", 
  return_tensors="pt")
sources = [
  tokenizer("The language of Spain is", 
    return_tensors="pt"),
  tokenizer("The capital of Italy is", 
    return_tensors="pt")]

intervened_outputs = parallel_gpt2(
  base, sources,
  {"sources->base": (
  # each list has a dimensionality of
  # [num_intervention, batch, num_unit]
  [[[1]],[[3]]],  [[[1]],[[3]]])}
)
\end{lstlisting} 
In the example above, we interchange the activations from the residual streams on top of the second token from the first example (``language'') as well as the fourth token from the second example (``Italy'') into the corresponding locations of the \texttt{base}'s computation graph. The motivating intuition is that now the next token might be mapped to a semantic space that is a mixture of two inputs in the \texttt{source} ``The language of Italy''. (And, in fact, ``Italian'' is among the top five returned logits.)

\subsection{Multi-Source Serial Interventions}

Interventions can also be sequentially applied, so that later interventions are applied to an intervened model created by the previous ones:
\begin{lstlisting}
serial_config = pv.IntervenableConfig([
  {"layer": 3, "component": "block_output"},
  {"layer": 10, "component": "block_output"}],
  # intervene on base one after another
  mode="serial")

serial_gpt2 = pv.IntervenableModel(
  serial_config, model=gpt2)

intervened_outputs = serial_gpt2(
  base, sources,
  # src_0 intervenes on src_1 position 1
  # src_1 intervenes on base position 4
  {"source_0->source_1": 1, 
   "source_1->base"    : 4}
)
\end{lstlisting} 
In the example above, we first take activations at the residual stream of the first token (``language'') at the 3rd layer from the first \texttt{source} example and swap them into the same location during the forward run of the second \texttt{source} example. We then take the activations of the 4th token (``is'') at layer~10 at upstream of this intervened model and swap them into the same location during the forward run of the \texttt{base} example. The motivating intuition is that the first intervention will result in the model retrieving ``The language of Italy'' and the second intervention will swap the retrieved answer into the output stream of the \texttt{base} example. (Once again, ``Italian'' is among the top five returned logits.)

\subsection{Intervenable Model}

The \texttt{IntervenableModel} class is the backend for decorating \texttt{torch} models with intervenable configurations and running intervened forward calls. It implements two types of hooks: \textbf{Getter} and \textbf{Setter} hooks to save and set activations.

Figure~\ref{fig:main} highlights \LibraryName{}'s support for LMs. Interventions can be applied to any position in the input prompt or any selected decoding step. 

The following involves a model with recurrent (GRU) cells where we intervene on two unrolled recurrent computation graphs at a time step:
\begin{lstlisting}
# built-in helper to get a GRU
_, _, gru = pv.create_gru_classifier(
  pv.GRUConfig(h_dim=32))
# wrap it with config
pv_gru = pv.IntervenableModel({
  "component": "cell_output",
  # intervening on time
  "unit": "t", 
  "intervention_type": pv.ZeroIntervention},
  model=gru)
# run an intervened forward pass
rand_b = torch.rand(1,10, gru.config.h_dim)
rand_s = torch.rand(1,10, gru.config.h_dim)
intervened_outputs = pv_gru(
  base = {"inputs_embeds": rand_b}, 
  sources = [{"inputs_embeds": rand_s}], 
  # intervening time step
  unit_locations={"sources->base": (6, 3)})
\end{lstlisting}  
A hook is triggered every time the corresponding model component is called. As a result, a vanilla hook-based approach, as in all previous libraries \citep{baukit,graphpatch,nnsight,mossing2024tdb}, fails to intervene on any recurrent or state-space model. To handle this limitation, \LibraryName{} records a state variable for each hook, and only executes a hook at the targeted time step.

\begin{figure*}[t!]
    \centering
    \includegraphics[width=0.32\textwidth]{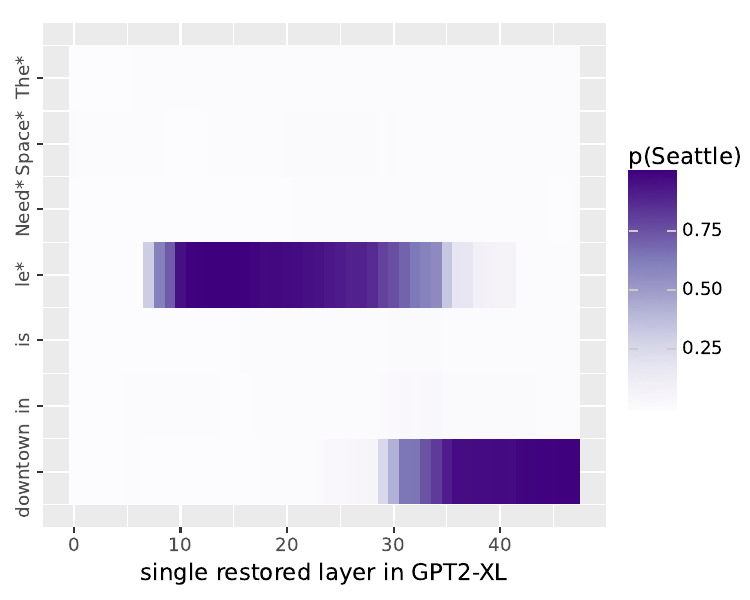}
    \includegraphics[width=0.32\textwidth]{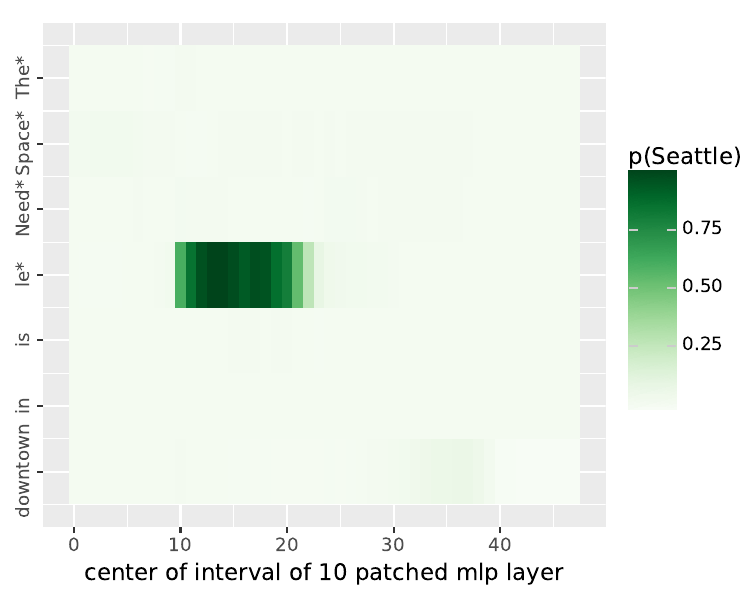}
    \includegraphics[width=0.32\textwidth]{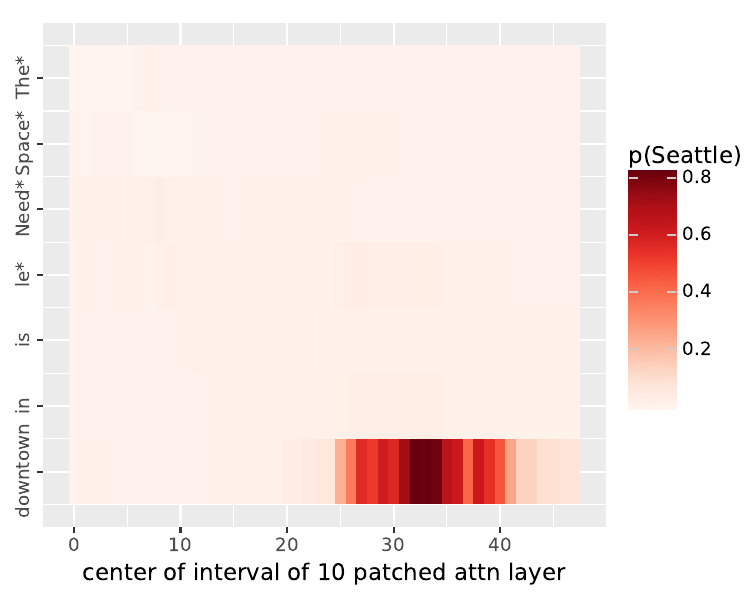}
    \caption{We reproduce the results in \citet{meng2022locating}'s Figure 1 of locating early sites and late sites of factual associations in GPT2-XL  \emph{in about 20 lines of \LibraryName{} code}. The causal impact on output probability is mapped for the effect of  each Transformer block output (left), MLP activations (middle), and attention layer output (right) .}
    \label{fig:three_figures}
\end{figure*}

\begin{figure}[t]
  \centering
  \includegraphics[width=1.0\linewidth]{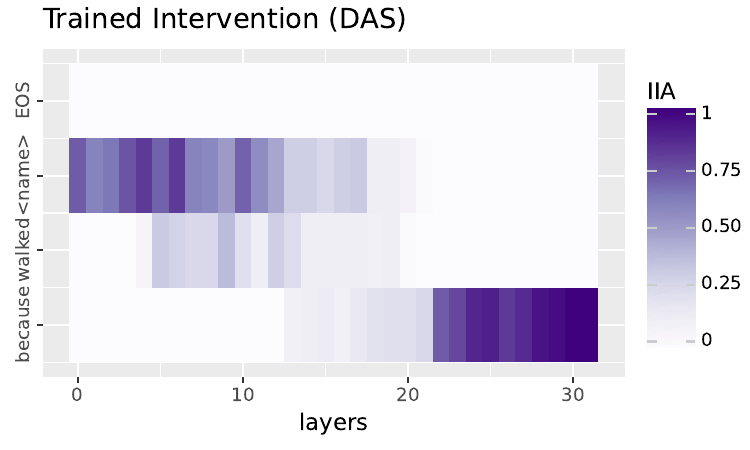}
  \includegraphics[width=1.0\linewidth]{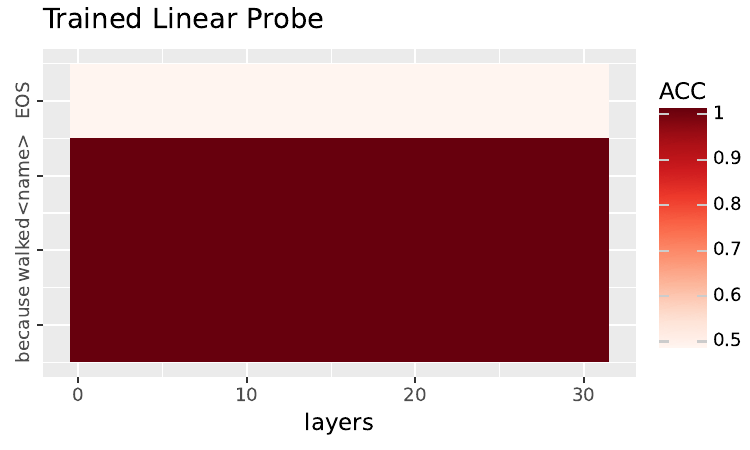}
  \caption{Results of interchange intervention accuracy (IIA) with the trainable intervention (DAS) and accuracy with the trainable linear probe on different model components when localizing gender information.}\label{fig:das_and_probe}
\end{figure}

\section{Case Study I: Locating Factual Associations in \texttt{GPT2-XL}}\label{sec:gpt2xl}
We replicate the main result in \citet{meng2022locating}'s Locating Factual Associations in \texttt{GPT2-XL} with \LibraryName{}. The task is to trace facts via interventions on fact-related datasets. Following \citeauthor{meng2022locating}'s setup, we first intervene on input embeddings by adding Gaussian noise. We then restore individual states to identify the information that restores the results. Specifically, we restore the Transformer block output, MLP activation, and attention output for each token at each layer. For MLP activation and attention output, we restore 10 sites centered around the intervening layer (clipping on the edges). Our \Figref{fig:three_figures} fully reproduces the main Figure 1 (p.~2) in \citeauthor{meng2022locating}'s paper. To replicate their experiments, we first define a configuration for causal tracing:
\begin{lstlisting}
def tracing_config(
    l, c="mlp_activation", w=10, tl=48):
  s = max(0, l - w // 2)
  e = min(tl, l - (-w // 2))
  config = IntervenableConfig(
    [{"component": "block_input"}] + 
    [{"layer": l, "component": c} 
      for l in range(s, e)],
    [pv.NoiseIntervention] +
    [pv.VanillaIntervention]*(e-s))
  return config
\end{lstlisting}  
With this configuration,  we corrupt the subject token and then restore selected internal activations
to their clean value. Our main experiment is implemented with about 20 lines of code with \LibraryName{}:
\begin{lstlisting}
trace_results = []
_, tokenizer, gpt = pv.create_gpt2("gpt2-xl")
base = tokenizer(
  "The Space Needle is in downtown", 
  return_tensors="pt")
for s in ["block_output", "mlp_activation", 
    "attention_output"]:
  for l in range(gpt.config.n_layer):
    for p in range(7):
      w = 1 if s == "block_output" else 10
      t_config, n_r = tracing_config(l, s, w)
      t_gpt = pv.IntervenableModel(t_config, gpt)
      _, outs = t_gpt(base, [None] + [base]*n_r,
        {"sources->base": ([None] + [[[p]]]*n_r,
          [[[0, 1, 2, 3]]] + [[[p]]]*n_r)})
      dist = pv.embed_to_distrib(gpt,
        outs.last_hidden_state, logits=False)
      trace_results.append(
        {"stream": s, "layer": l, "pos": p, 
         "prob": dist[0][-1][7312]})
\end{lstlisting}

\section{Case Study II: Intervention and Probe Training with \texttt{Pythia-6.9B}}
We showcase intervention and probe training with \LibraryName{} using a simple gendered pronoun prediction task in which we try to localize gender in hidden representations. For trainable intervention, we use a one-dimensional Distributed Alignment Search (DAS; \citealp{geiger2023finding}), that is, we seek to learn a 1D subspace representing gender. To localize gender, we feed prompts constructed from a template of the form ``[\textbf{John}/\textbf{Sarah}] walked because [\textbf{he}/\textbf{she}]'' (a fixed length of 4) where the name is sampled from a vocabulary of 47 typically male and 10 typically female names followed by the associated gendered pronoun as the output token. We use \texttt{pythia-6.9B} \citep{biderman2023pythia} in this experiment, which achieves 100\% accuracy on the task. We then train our interventions and probes at the Transformer block output at each layer and token position. For intervention training, we construct pairs of examples and train the intervention to match the desired counterfactual output (i.e., if we swap activations from an example with a male name into another example with a female name, the desired counterfactual output should be ``he''). For linear probe training, we use activation collection intervention to retrieve activations to predict the pronoun gender with a linear layer. 

As shown in \Figref{fig:das_and_probe}, a trainable intervention finds sparser gender representations across layers and positions, whereas a linear probe achieves 100\% classification accuracy for almost all components. This shows that a probe may achieve high performance even on representations that are not causally relevant for the task.

\section{Limitations and Future Work}
We are currently focused on two main areas:
\begin{enumerate}[itemsep=0pt, topsep=4pt]
  \item 
Expanding the default intervention types and model types. Although \LibraryName{} is extensible to other types, having more built-in types helps us to onboard new users easily. 
  \item \LibraryName{} is designed to support complex intervention schemes, but this comes at the cost of computational efficiency. As language models get larger, we would like to investigate how to scale intervention efficiency with multi-node and multi-GPU training.
\end{enumerate}

\section{Conclusion}
We introduce \LibraryName{}, an open-source \texttt{Python} library that supports intervention-based research on neural models. \LibraryName{} supports customizable interventions with complex intervention schemes as well as different families of model architectures, and intervened models are shareable with others through online model hubs such as HuggingFace. Our hope is that \LibraryName{} can be a powerful tool for discovering new ways in which interventions can help us explain and improve models.

\bibliography{custom}

\end{document}